# Light Field Spatial Resolution Enhancement Framework

Javeria Shabbir‡, *Muhammad Zeshan. Alam*†, *and M. Umair Mukati* ∗

‡*Georgia Institute of Technology,* †*Department of Computer Science, Brandon University, MB, Canada*
*DTU Electro, Technical University of Denmark, 2800 Kgs. Lyngby, Denmark
(corresponding author e-mail: alamz@brandonu.ca)

*Abstract*—Light field (LF) imaging captures both angular and spatial light distributions, enabling advanced photographic techniques. However, micro-lens array (MLA)-based cameras face a spatial-angular resolution trade-off due to a single shared sensor. We propose a novel light field framework for resolution enhancement, employing a modular approach. The first module generates a high-resolution, all-in-focus image. The second module, a texture transformer network, enhances the resolution of each light field perspective independently using the output of the first module as a reference image. The final module leverages light field regularity to jointly improve resolution across all LF image perspectives. Our approach demonstrates superior performance to existing methods in both qualitative and quantitative evaluations.

I. INTRODUCTION

Contrary to traditional cameras, the light field (LF) camera captures the light rays approaching its surface, preserving the angular information of the incident light rays on the camera's sensor [1], [2]. This directional information enables a variety of new applications in computer vision and image processing. The features of light field imaging include post-capture refocusing, synthetically varying aperture size, 3D rendering and depth estimation. These additional capabilities are achieved by separately recording intensities of light rays from different directions at each pixel position [3], [4]. In light field parameterization, ray positions and directions are typically represented in 3D and 2D coordinates, respectively, along with the light's physical properties (e.g., wavelength, polarization). However, for practical applications where energy loss is negligible and only light intensity is considered, this model simplifies to a four-dimensional space [5], [6]. This approach is widely adopted in practical light field acquisition methods described below.

Light field acquisition can be done in a many ways including micro-lens arrays (MLAs) [7]–[9], coded masks [10]–[12], camera arrays [13]–[15], and among these different implementations, MLA-based light field cameras offer a cost-effective approach, leading to commercial LF cameras namely Lytro and Raytrix [7], [8]. The main drawback of MLA-based light field cameras is the fundamental spatial-angular resolution trade-off [16] since a single sensor is shared to record both angular and spatial information. For example, the first-generation Lytro camera has a sensor of around 11 megapixels, producing 11x11 angular resolution and less than 0.15 megapixel spatial resolution, significantly lower than current standards [17].

To overcome the spatio-temporal resolution trade-off, several methods have been developed [18]–[20], [20]–[25]. In [18] variational super-resolution technique was introduced. Hybrid camera designs were proposed in [24] and [25], and the more recent deep learning-based models are presesnted in [19]–[22], and [23]. In [19], authors use a learning-based approach that utilizes a residual convolutional network to achieve high spatial resolution. Whereas, [20] proposes to incorporate information from different views via a graph regularizer to improve spatial resolution of the whole light field. [21] incorporates contextual information from multiple scales into bidirectional recurrent CNN which uses spatial information from neighboring patches of light field. [22] explored a method that super-resolves each sub-aperture image of the light field individually and then in order to maintain parallax structure, a regularization network was appended. State-of-the-art method proposed by [23] extracts angular and spatial information separately and later combines them interactively to super-resolve each patch of light field indicating that both angular and spatial information is necessary for the task of super-resolution.

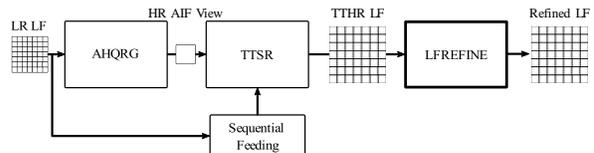

Fig. 1: Block diagram of proposed method.

In this paper, we propose a modular framework to improve the LF spatial resolution, as highlighted in Figure 1. Specifically, we generate a high resolution version of the central perspective image of the low-resolution input LF using an All-In-Focus High Quality Reference Generator (AHQRG) network. AHQRG's output is fed into the Texture Transformer Network for Image Super-Resolutionutilize (TTSR) as a high-quality reference image to improve the resolution of all the views of the light field independently. We further exploit the regularity in light field images by the so-called Light field refinement (LFREFINE) network to jointly enhance the resolution of all the LF perspective images.

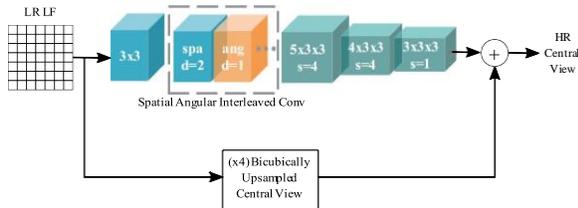

Fig. 2: Block diagram of All-In-Focus High Quality Reference Generator (AHQRG).

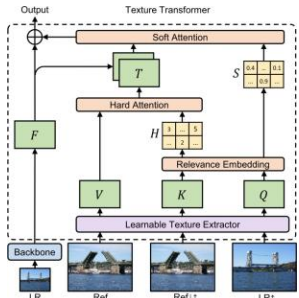

Fig. 3: Texture Transformer Network (TTSR).

## II. PROPOSED LF RESOLUTION ENHANCEMENT FRAMEWORK

To enhance the spatial resolution of light-field images fourfold, we introduce a structured methodology comprising three distinct modules as follows.

### A. All-In-Focus High-Quality Ref. Generator

The AHQRG module is pivotal for generating a high-resolution, all-in-focus reference image from a low-resolution light field input. This is achieved through a sophisticated network architecture that incorporates spatial-angular interleaved convolution, as shown in Figure2.
The design intricately processes $7 \times 7$ views of the input light field, converging them into a singular, enhanced output. Key to this process is a series of initial $3 \times 3$ convolutional layers followed by interleaved filtering that adeptly balances between angular and spatial data dimensions. Subsequent three 3d convolution layers are dedicated to extracting and integrating information across both spatial and angular domains, culminating in a residual addition to a bicubically upsampled central view, thus ensuring the production of a high-quality central image.

### B. Texture Transformer Network

The Texture Transformer Super-Resolution (TTSR) module is designed to enhance the resolution of low-resolution input images by leveraging the textural information present in a high-resolution reference image. This network is adapted in our work to address the super-resolution of individual sub-aperture images within a light field. The TTSR module operates by receiving two primary inputs: a low-resolution view from the light field and a high-resolution reference image generated by the AHQRG module. It produces a high-resolution version of the corresponding low-resolution light field view.

As depicted in Figure 3 the TTSR module employs a texture transformer that extracts features from the high-resolution reference image to enhance the low-resolution input. The process involves using a bicubically down-sampled and then up-sampled version of the reference image as the 'key', the bicubically up-scaled low-resolution image as the 'query', and the original high-resolution reference image as the 'value' in the texture transformer framework. This method enables the effective transfer of high-resolution textures from the reference image to the low-resolution target, significantly improving the quality of the super-resolved output.

### C. LF Refinement

The TTSR module responsible for the resolution enhancement of individual views within a light field, while effective, may inadvertently introduce irregularities that deviate from the inherent structural coherence typical of light fields. Recognizing this potential discrepancy not as a limitation but as an opportunity for further refinement, our methodology incorporates an additional step designed to preserve the light field's structural integrity.
To address this challenge and enhance the overall quality of the super-resolved light field, we introduce a novel phase in our processing pipeline: the light field blending step. This crucial enhancement employs an Epipolar Plane Image (EPI) loss function, as described in [26], serves a dual purpose. Firstly, it enforces the light field prior, ensuring that the super-resolved views adhere to the expected regularity and continuity of the light field structure. Secondly, it acts as a mechanism for quality improvement, leveraging the spatial and angular consistency inherent to light fields to guide the super-resolution process towards outputs that not only exhibit high resolution but also maintain the characteristic coherence of light field imagery.

## III. DATASET PREPARATION AND TRAINING

For our experiments, the dataset from Kalantari et al. [27] was utilized. Each LF has an original angular resolution of $8 \times 8$, out of which we select $7 \times 7$ top-left views, exclusively using the luminance channel for simplicity. To enhance training robustness and adaptability to varied lighting conditions, we applied gamma correction with coefficients ranging from 0.4 to 1.0 as a data augmentation technique. The training involved using randomly cropped $128 \times 128$ patches from the light fields, which were down-scaled bicubically to $32 \times 32$ for generating low-resolution inputs.

The AHQRG module processes a $7 \times 7 \times 32 \times 32$ low-resolution light field, outputting a single $128 \times 128$ spatial resolution view. This All-In-Focus (AIF) output, necessary for subsequent Texture Transformer Super-Resolution (TTSR) processing, assumes an inherently wide depth-of-field characteristic of each light field view, justifying the use of the unscaled $128 \times 128$ central view as a training target. The central view's strategic selection is due to its advantageous position, potentially integrating information from surrounding views.

In our proposed framework, the AHQRG and LFREFINE modules undergo training, leveraging a pre-trained TTSR module to enhance efficiency and effectiveness. We used Adam Optimizer, adhering to PyTorch's default beta values. A learning rate of $10^{-4}$ was uniformly applied to these modules, with their architecture and hyper-parameters.

The AHQRG module's loss function integrates the sum of L1 losses for the light field and its gradient, emphasizing accuracy and gradient fidelity. The LFREFINE module adopts the AHQRG's loss framework while incorporating an Epipolar Plane Image (EPI) gradient loss to reinforce the structural integrity of the light field. Training of the AHQRG and LFREFINE modules persisted until the models demonstrated convergence, a milestone achieved at approximately 120 and 250 epochs, respectively. This iterative training process ensured that each module was optimized for performance, contributing to the overall efficacy of our proposed scheme in light field super-resolution and refinement.

## IV. EXPERIMENTS AND RESULTS

In this section, we provide the quantitative and qualitative evaluation of the proposed LF resolution enhancement framework. In Figure 4 it can be observed that significant gain in PSNR is achieved over bicubically resized LF for the TTSR that operates on the output of the AHQRG module as reference. Although, TTSR is able to generate a high-quality central perspective, but as soon as the perspective changes from the central location the quality starts dropping significantly. We believe that this may be due to incorrect placement of texture. On the other hand, LFREFINE considerably improves the performance by applying light field prior. Not only does this module improve the quality of views away from the central location but also improves the quality of central view.

Next, we qualitatively compare the quality of the output of each of the modules and the bicubically resized image. From Figure 5, a significant gain can be observed in the quality of the central views over the bicubically resized image. Furthermore, it may be seen that for some examples, the quality of the reference image is higher than the outputs of TTSR and LFREFINE modules. It may be the case that TTSR replaces the actual texture with a different texture.

Figure 6 further compares the performance at the top-left view index of the light field. It may be seen that LFREFINE module robustly improves the quality of the light field generated by TTSR, showing the effectiveness of using the refinement step.

## V. CONCLUSION

We propose a modular framework to enhance the quality of a LF. The proposed scheme is composed of three sub-modules that have well-defined roles in the generation of a high-quality LF. The qualitative comparison demonstrates the superiority of the proposed framework over TTSR method. Quantitatively, we have shown that the proposed scheme achieves around 4 to 6 dB improvement in PSNR over bicubically resized LF.

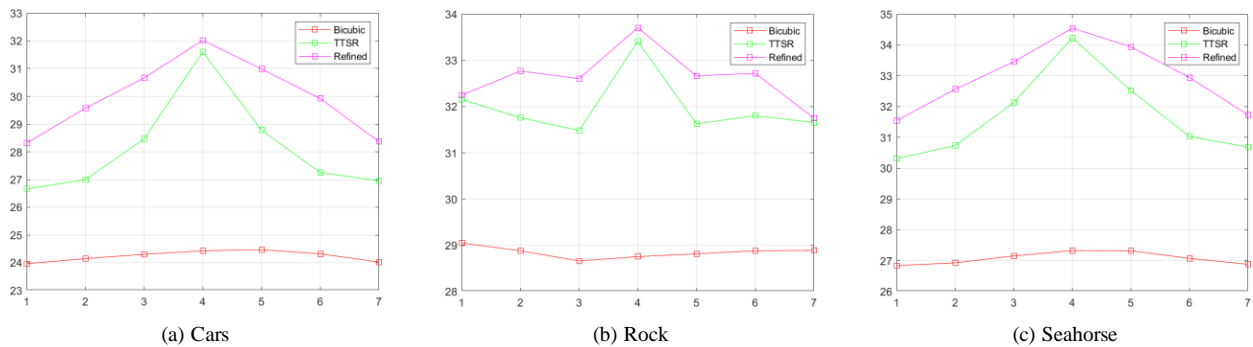

(a) Cars  (b) Rock  (c) Seahorse

Fig. 4: Plotting the PSNR of the diagonal views of the light fields for bicubically resized view, and outputs from TTSR and LFREFINE.

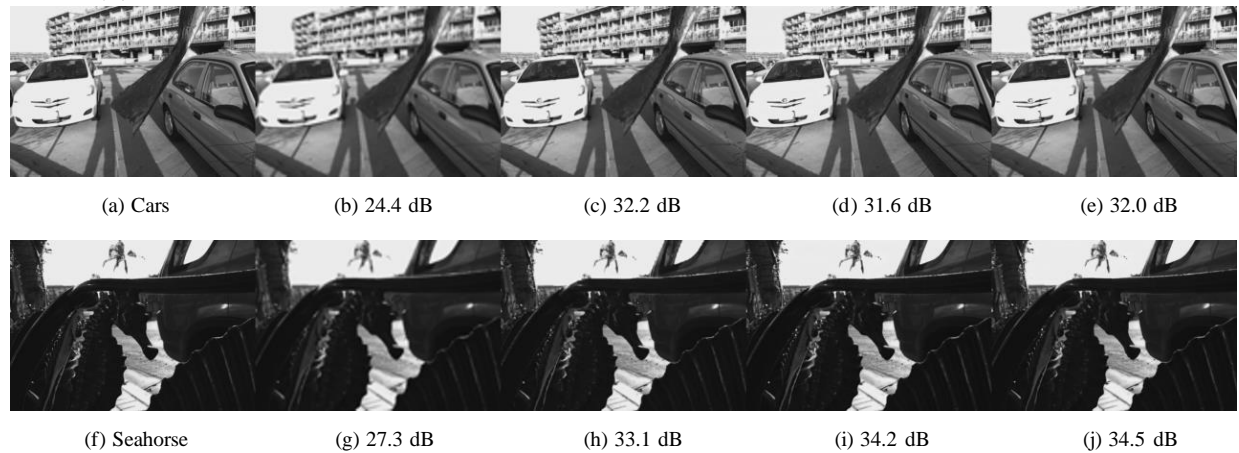

(a) Cars  (b) 24.4 dB  (c) 32.2 dB  (d) 31.6 dB  (e) 32.0 dB

(f) Seahorse  (g) 27.3 dB  (h) 33.1 dB  (i) 34.2 dB  (j) 34.5 dB

Fig. 5: Visual demonstration of reconstructed central views at different stages of the proposed scheme. (Left to Right) Ground truth, Bicubic, AHQRG, TTSR, LFREFINE.

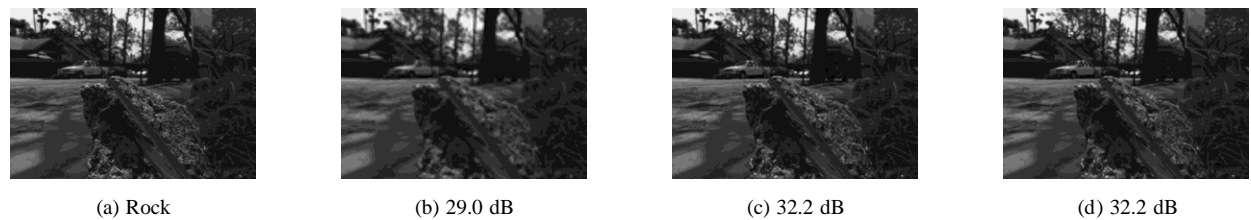

(a) Rock  (b) 29.0 dB  (c) 32.2 dB  (d) 32.2 dB

Fig. 6: Top-left view at different stages of the proposed scheme. (Left to Right) Ground truth, Bicubic, TTSR, LFREFINE.